\documentclass{llncs}
\usepackage{makeidx}  

\usepackage{amsmath}
\usepackage{amsfonts}
\usepackage[colorinlistoftodos]{todonotes}

\begin{document}
%
%
\title{Aggregation of binary feature descriptors for compact scene model representation in large scale structure-from-motion applications}
\titlerunning{Aggregation of binary feature descriptors}  
%
\author{Jacek Komorowski\inst{1} \and Tomasz Trzci{\'{n}}ski\inst{2}}

\authorrunning{J. Komorowski, T. Trzci{\'{n}}ski}   

\tocauthor{Jacek Komorowski, Tomasz Trzci{\'{n}}ski}

\institute{Warsaw University of Technology, Warsaw, Poland,\\
\email{jacek.komorowski@gmail.com}
\and
Warsaw University of Technology, Warsaw, Poland,\\
\email{t.trzcinski@ii.pw.edu.pl}}

\maketitle              

\begin{abstract}
In this paper we present an efficient method for aggregating binary feature descriptors
to allow compact representation of 3D scene model in incremental structure-from-motion and SLAM applications.
All feature descriptors linked with one 3D scene point or landmark are represented by a single low-dimensional real-valued vector called a \emph{prototype}.
The method allows significant reduction of memory required to store and process feature descriptors in large-scale structure-from-motion applications.
An efficient approximate nearest neighbours search methods suited for real-valued descriptors, such as FLANN~\cite{flann1}, can be used on the resulting prototypes to speed up matching processed frames.
\end{abstract}
\section{Introduction}

In recent years a number of methods was published aimed at constructing 3D point cloud models from video sequences or large collections of images using structure-from-motion techniques~\cite{snavely1,agarwal1,frahm1}. 
In a typical incremental structure-from-motion pipeline~\cite{sch1} keypoints are detected on each image in a video sequence and local feature descriptors are computed. 
Feature descriptors in different images are matched which allows estimating camera poses. 
3D scene model, in the form of a sparse point cloud, is constructed by triangulating matching feature descriptors.
Position of reconstructed 3D points and camera poses are iteratively refined using bundle adjustment~\cite{triggs1999bundle} method.
For each 3D scene point, so called landmark, a list of feature descriptors used to construct it is kept. This is necessary to allow recognizing previously visited places and for loop closure.
When a new frame is processed, feature descriptors detected on the frame are matched with descriptors linked with previously reconstructed landmarks. This is usually done by searching for the nearest descriptor (in the descriptor space, Hamming for binary descriptors or Euclidean for real-valued descriptors) and taking the corresponding landmark.
Such 2D (feature descriptor) to 3D (landmark) correspondence is noisy and contains a large number of incorrect matches. Robust parameter estimation methods such as RANSAC~\cite{fischler1987random} are used to estimate the absolute pose of the new video frame with respect to the 3D scene model.

In large scale structure from motion applications tens of thousands or millions~\cite{heinly2015} of images are processed and few hundred local feature descriptors are usually detected on each processed image. 
This produces very large datasets consisting of millions or hundreds of millions of feature descriptors. 
Although the size of an individual feature descriptor is small, storing millions of descriptors requires significant storage which can be problematic on mobile devices.

Floating-point descriptors, such as SIFT~\cite{lowe2004}, typically offer better performance but at a higher computational cost. Their binary competitors, such as FREAK~\cite{alahi1}, are significantly faster to compute.
In this work we focus our attention on feature binary descriptors. Due to their computational efficiency, they are often chosen in practical structure-from-motion applications on mobile devices with limited hardware resources.

In this paper we propose an efficient method 
for compressing scene models in incremental structure-from-motion applications
by aggregating binary feature descriptors linked with each reconstructed landmark.
Our approach is based on the the idea of computing a compact \emph{prototype} 
representing all binary feature descriptors linked with each landmark.
Only this compact \emph{prototype} is stored and used in further processing to match feature descriptors from subsequent frames.
Original binary feature descriptors are discarded to free up the memory.
The \emph{prototypes} are iteratively updated, as new feature descriptors from subsequent frames are matched with corresponding landmarks.
The method allows significant reduction of memory required to store feature descriptors from previously processed frames and speeds up matching feature descriptors from new frames.
Our method was inspired by prototypical networks~\cite{snell1} which were proposed for few-shot learning domain.

\section{Related work}

Due to an increasing number of practical applications of large scale structure-from-motion and SLAM methods, a number of research papers was published on efficient storage and compression of 3D scene models.

One approach is to compress large scene maps constructed in a structure-from-motion application by selecting representative 3D scene points (landmarks).
\cite{li1} proposes a method to compress 3D scene model by storing and processing only a small subset of representative landmarks.
The method selects a minimal set of landmarks that sufficiently cover the scene visible on processed images.
The problem is formulated as a set covering problem and the smallest subset of points covering all processed images is selected.
A greedy algorithm finding an approximate solution is proposed.
\cite{cao1} uses more sophisticate approach to produce the reduced scene description. The method selects landmarks taking into account their distinctiveness and coverage of the scene. A greedy algorithm is used to incrementally create a compact subset of landmarks. The method tries to balance two goals: maximize the probability of registering a new image while minimizing the number of points selected. Authors claim they can summarize a structure-from-motion model with as little as 3\% of original landmarks while keeping reasonable image registration performance.
Above methods differ from our approach, as they aim at selecting representative landmarks (3D points) that can be used to reliably match subsequent frames. Our method aggregates feature descriptors linked with an individual landmark. It is complementary to these methods  and can be used in conjunction to further reduce the memory required to store 3D scene model.

Efficient compression methods are proposed for real-valued feature descriptors such as SIFT.
Mean-shift clustering is used in~\cite{irschara1} to compress SIFT descriptors linked with each landmark. 
Authors report 50\% reduction in memory footprint without adversely affecting matching performance.
However this method is not directly applicably to binary feature descriptors as clustering-based methods perform poorly in binary spaces~\cite{trzcinski}.
\cite{Johnson1} proposes a general feature descriptor compression method using tree coding technique. The method allows fast descriptor matching without requiring decompression. However it's aimed at real-valued descriptors such as SIFT, SURF or GLOH.

\cite{Opden,redondi2013,baroffio2014} propose compression methods aimed at minimizing the bandwidth required to transmit feature descriptors. 
These methods can be used for efficient transmission of feature descriptors and are not directly applicable in a structure-from-motion reconstruction pipeline.


A few methods~\cite{7574762,amato2018} try to solve the solve the problem of a compact image representation by aggregating local feature descriptors in the image.
Quantization or aggregation techniques are used to generate summarization of all the extracted
features in one image. This allows representing an image by a single global descriptor rather
than hundreds of local feature  descriptors. 
Our method, in contrast, tries to solve a different problem -- how to compactly represent similar feature descriptors from multiple images linked with a single landmark.



\cite{kom1} investigates if data-dependent hashing methods can be applied to find more compact representation of binary feature descriptors.
A  representative sample of recent unsupervised, semi-supervised and supervised hashing methods is experimentally evaluated on large datasets of labelled binary FREAK feature descriptors. 
The results prove that hashing methods cannot be effectively used to find compact representation of binary feature descriptors without sacrificing nearest neighbours search precision.

\section{Method}
\label{jk:section:method}

This section presents a  method to efficiently compress a scene map
constructed in large-scale incremental structure-from-motion applications.
It is based on the simple idea of calculating a compact representation, so called \emph{prototype}, of all descriptors linked with each landmark (3D scene point).
Feature descriptors used to construct the \emph{prototype} are discarded and removed from the memory.
The \emph{prototype} is then used to match feature descriptors from subsequent frames.
During incremental processing of the video sequence, the \emph{prototype} is updated when new feature descriptors detected on new frames, are linked with the corresponding landmark.
Suppose, in incremental structure-from-motion pipeline we initially construct the landmark by triangulating three keypoints detected in three frames. 
\emph{Prototype} is initially computed using these three corresponding feature descriptors.  
Then, when new keypoint from some subsequent frame is linked with the landmark, its \emph{prototype} is updated to take into account the new feature descriptor. 

The simplest approach would be to compute the \emph{prototype} as a mean of feature descriptors linked with one landmark. Then, the mean will be iteratively updated, when new descriptors, from subsequently processed frames, are linked with the landmark. 
However, when using binary feature descriptors, such approach is not feasible.
Calculating and storing the real-valued mean of binary descriptors will increase memory requirements multiple times. For each bit of a binary descriptor we would need to store its real-valued mean. 
Alternatively, we can calculate mean value of each bit and threshold it at 0.5, to produce a binary-valued quantized mean vector. But this approach does not allow interactive updating. 
If a new feature descriptor in some subsequent frame is linked with the landmark, we would not known how to update the quantized mean with values of the new feature descriptor. 
\footnote{Suppose the first bit of a quantized mean is 0 and first bit of a new feature descriptor is 1. As original feature descriptors used to construct the quantized mean are discarded, we cannot determine if the first bit in an updated quantized mean should remain 0 or be changed to 1.}
Therefore such approach cannot be used in an incremental structure-from-motion processing pipeline.

Our method works by using a neural network to compute low dimensional, real-valued embedding of binary feature descriptors. 
Then, the \emph{prototype} is calculated as an arithmetic mean of real-valued embeddings of all descriptors linked with one landmark.
This produces the low dimensional, real-valued \emph{prototype} representing all descriptors linked with a landmark.
We can then discard  values of all feature descriptors used to calculate the \emph{prototype}, freeing up the memory.
The only additional data that needs to be stored for each landmark, in addition to the \emph{prototype}, is the number of feature descriptors used to compute the mean. This allows updating the \emph{prototype} as feature descriptors linked with the same landmark are detected on new frames.
Usually the number of descriptors linked with a landmark is not bigger than 255, so one byte per landmark is sufficient.


\begin{figure}
\centering
\label{jk:method}
\includegraphics[width=0.7\textwidth, trim={0.1cm 0.1cm 0.1cm 0.1cm},clip]{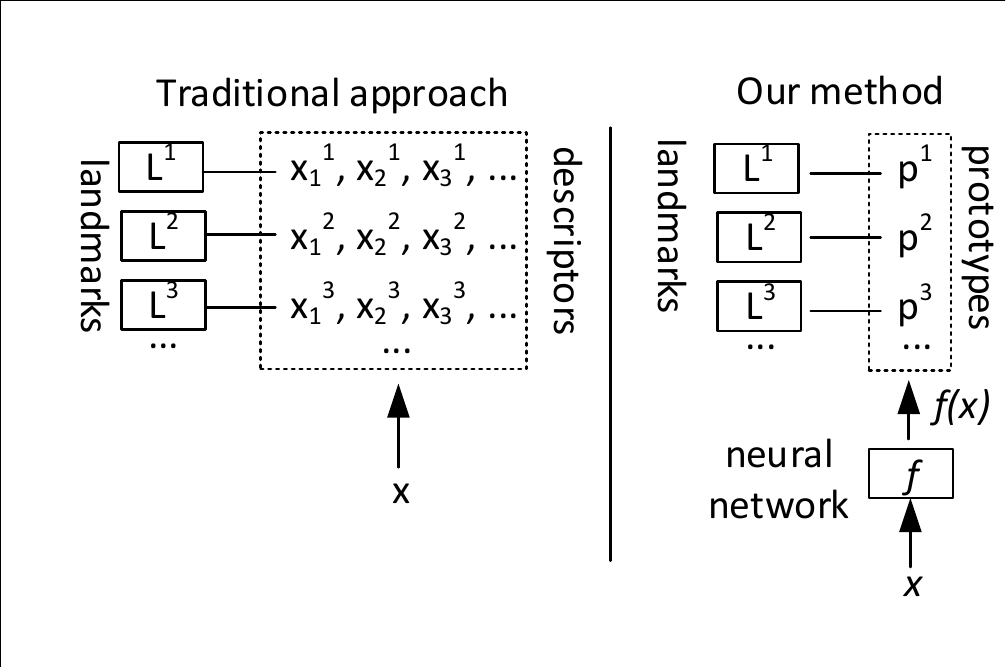}
\caption{(Left) In a traditional approach a feature descriptor $\mathbf{x}$ detected on a new frame is matched with all features linked with reconstructed landmarks.
In our method (Right) an embedding $f(\mathbf{x})$ of a feature descriptor is computed using the neural network $f$. Then the embedding is matched with prototypes linked with reconstructed landmarks. Single prototype per each landmark is stored.} 
\end{figure}

Formally, let $\mathbf{p}_i \in \mathbb{R}^k $ denote a $k$-dimensional prototype linked with $i$-th landmark (3D scene point).
$n_i$ is a number of feature descriptors used to construct the prototype $\mathbf{p_i}$.
$\mathbf{x} \in \left\{ 0,1 \right\} ^{b}$ is a $b$-dimensional binary feature descriptor ($b=512$ for FREAK) and 
$\mathbf{y} = f(\mathbf{\mathbf{x}}) \in \mathbb{R}^k$ is a low-dimensional, real valued, embedding of a feature descriptor $\mathbf{x}$ computed by a the neural network.
Initially a \emph{prototype} $\mathbf{p}_i$ is computed as a mean of real-valued embeddings of feature descriptors used to construct the $i$-th landmark:
\begin{equation}
\label{jk:eq1}
  \mathbf{p}_i = \frac{1}{n_i} \sum_{j=1}^{n_i} f \left( \mathbf{\mathbf{x}} \right) \ .
\end{equation}
When a new frame is processed, for each feature descriptor $\mathbf{x}$ detected on the frame, its real-valued embedding $\mathbf{y} = f ( \textbf{x} )$ is computed. Then we use the embedding $\mathbf{y}$ to search for the nearest neighbour, in Euclidean distance sense, in a set of landmark prototypes.
The landmark linked with the closest, in Euclidean distance sense, prototype is retained as a putative match. The search can be efficiently done using fast approximate nearest neighbour search method such as FLANN~\cite{flann1}.
Next, the robust parameter estimation method, such as RANSAC~\cite{fischler1987random} is used to estimate the absolute pose and orientation of the new video frame with respect to the 3D model and to filter out putative matches.
When a new keypoint is linked with the $i$-th landmark, its \emph{prototype} is updated using the formula:
\begin{equation}
\label{jk:eq2}
\mathbf{p}_i' = \frac{n_i}{n_i+1} \mathbf{p}_i + \frac{1}{n_i+1} f \left( \mathbf{\mathbf{x}} \right) \ .
\end{equation}
This calculates the new prototype as a weighted average of the previous prototype and embedding of the new feature descriptor.
See Fig. \ref{jk:method} for visualization of our method.

In practice, in addition to  prototypes linked with each landmark, we need to keep all feature descriptors detected in few previously processed frames. This allows constructing new landmarks by triangulating matching feature descriptors in subsequent frames and provides initialization to our method.


We consider two approaches to train the neural network to compute low-dimensional embeddings of binary feature descriptors:
Triplet Networks \cite{hoffer1} and Prototypical Networks \cite{snell1}.

\begin{figure}
\centering
\label{jk:triplet_network}
\includegraphics[width=0.7\textwidth, trim={0.1cm 0.1cm 0.1cm 0.1cm},clip]{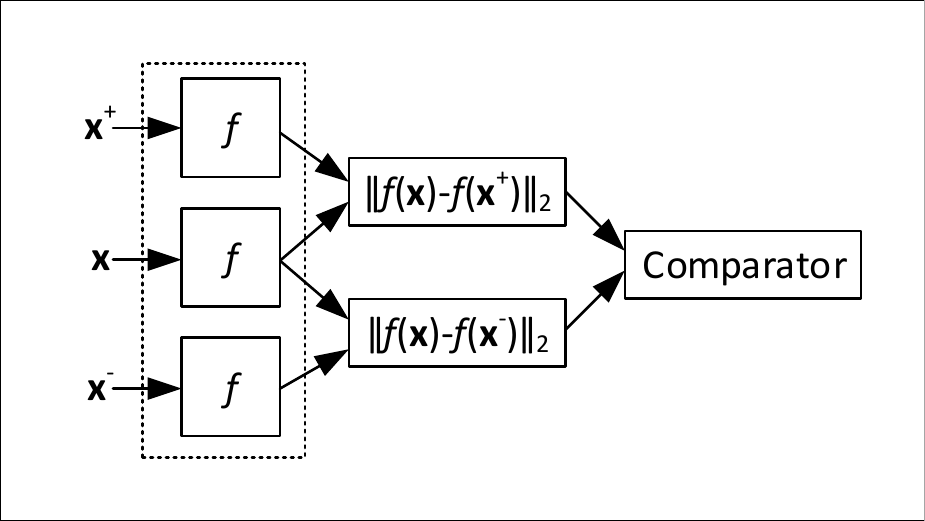}
\caption{High-level architecture of the triplet network~\cite{hoffer1}.}
\end{figure}

\paragraph{Triplet networks} \cite{hoffer1} consist of three identical modules with shared weights that compute embeddings of the input data.
High-level architecture of Triplet neural network is depicted on Fig. \ref{jk:triplet_network}.
During the training triplets of elements are presented to the network. 
The triplets are in the form of $(\mathbf{x}, \mathbf{x^+}, \mathbf{x^-})$, where $\mathbf{x}$ in an anchor element randomly sampled from a training set, $\mathbf{x^+}$ is a positive example from the same class and $\mathbf{x^-}$ is a negative example from a different class.

Input elements in a triplet are processed by an embedding module $f$ which computes their embeddings 
 $f(\mathbf{x})$, $f(\mathbf{x}^+)$ and $f(\mathbf{x}^-)$.
The embeddings are used to compute the triplet loss, as formulated in \cite{balntas1}:
\begin{equation}
\label{jk:eq4}
L(\mathbf{w}) = \frac{1}{N} \sum_{i=1}^N \max \{
d_{\mathbf{w}}( \mathbf{x}_i, \mathbf{x}_i^+) - d_{\mathbf{w}}(\mathbf{x}_i, \mathbf{x}_i^- ) + {\rm margin}, 0\}  ,
\end{equation}
where \( d_{\mathbf{w}}(\mathbf{x}, \mathbf{y}) = \left\lVert f_{\mathbf{w}} ({\bf x}) - f_{\mathbf{w}} ({\bf y}) \right\rVert_2 \) is an Euclidean distance between learned embeddings and $f_{\mathbf{w}}$ a function computed by an embedding module of the Triplet network parametrized by a weight vector $\mathbf{w}$.
    
The training is done using a classical mini-batch gradient descent approach.
The training procedure optimizes the weights of the network, so that for each triplet the distance from an embedding of an anchor element $\mathbf{x}$ to the embedding of a negative example $\mathbf{x}^{-}$ is greater by a margin than a distance to an embedding of a positive example $\mathbf{x}^{+}$.
%
%

\paragraph{Prototypical networks} \cite{snell1} work differently, as their loss function models directly results of the nearest neighbour search in a prototype space.
Prototypical networks compute $M$-dimensional representation $\mathbf{c}_k \in \mathbb{R}^M$, called \emph{prototype}, for each class using an embedding function $f_{\mathbf{w}} : \mathbb{R}^N \rightarrow \mathbb{R}^M$ 
parametrized by a weight vector $\mathbf{w}$.
Training episodes are formed by randomly selecting a subset of classes from the training set. Then a subset of examples from each class forms a support set and the remaining ones serve as query points.
Prototypical networks produce a distribution \( p_\mathbf{w} \) over classes for each query point \( \mathbf{x} \) based on a softmax over distances to the prototypes build using embeddings of elements from the support set.
\begin{equation}
\label{jk:eq3}
p_\mathbf{w}(y = k | \mathbf{x}) 
=
\frac{\exp \left(  -d \left( f_{\mathbf{w}} \left( \mathbf{x} \right), \mathbf{c}_k   \right) \right)}
{\sum_{k'} \exp \left(  -d \left( f_{\mathbf{w}} \left( \mathbf{x} \right), \mathbf{c}_{k'}   \right) \right)} ,
\end{equation}
where 
\( \mathbf{c}_k \) is a \emph{prototype} for each class calculated as a mean vector of the embedded support points from the class:
\(
\mathbf{c}_k = 
\sum_{ \mathbf{x}_i \in S_k }
f_{\mathbf{w}} \left( \mathbf{x}_i \right)
/
\left| S_k \right|
\) and \( S_k \) is a support set for class \( k \).
The loss \( L \) minimized during the training using stochastic gradient descent is defined as the mean negative log-probability of the true class $k$ for all elements from the query set:
\(
L(\mathbf{w}) 
= 
\sum_k
\sum_{\mathbf{x} \in Q_{k}}
- \log p_\mathbf{w}(y = k | \mathbf{x})
\),
where \( Q_k \) is a query set for class \( k \).
For details of the training procedure see \cite{snell1}.

\section{Experimental evaluation}
\label{jk:section:experimental}
%

%
%

\paragraph{Datasets}

All experiments are conducted using data acquired with structure-from-motion solution embedded in a Google Tango tablet. The device produces datasets containing keypoints and feature descriptors detected in the input video sequence. Camera poses and scene structure, in the form of sparse 3D point set, are reconstructed using reliable structure-from-motion methods. They are used as the ground truth during the network training and performance evaluation. 
The training sequence consists of over 4 million  FREAK descriptors detected in almost 10 thousand keyframes.
The validation sequence, used to choose the best network architecture, consists of almost 2 million  FREAK descriptors detected in approximately 5 thousand keyframes. 
The test sequence, used to measure the final performance of our method, consists of over 2 million  FREAK descriptors detected in approximately 5 thousand keyframes. 

%
%

\paragraph{Performance metric}

In order to choose the best embedding network architecture and evaluate performance of the resulting prototypes we simulated one step in a typical structure-from-motion pipeline.
The evaluation set is split into two parts. 
90\% of keyframes are randomly chosen and form the support set. They simulate frames already seen and processed in a structure-from-motion processing pipeline. 
Based on the ground truth data, we compute embeddings of all descriptors linked with each landmark using the network being evaluated and compute prototypes using Eq. \ref{jk:eq1}.
Then unaggregated feature descriptors are discarded and only one computed prototype per each landmark is retained.
Remaining 10\% of keyframes form a query set. They simulate new frames that are being matched to the scene model reconstructed using previously seen frames.
We randomly sample 10 thousand descriptors from the query subset. 
For each sampled descriptor we compute its embedding using the network being evaluated and find the closest, in Euclidean distance sense, prototype in the support set.
If the prototype found is linked with the same landmark as the sampled descriptor, we declare a correct match otherwise we have a failure.
The \emph{precision} is defined as a proportion of correct matches to all search attempts.
The procedure described above simulates matching descriptors from new frames with landmarks in a 3D model of the scene constructed using previously processed frames.

%
%

\paragraph{Network training}

We systematically evaluate wide range of neural network architectures to find the best performing one.
We expect that prototype networks yield the best performance, as the their loss function directly models results of the nearest neighbour search in a prototype space.
Loss function for Triplet networks aims at computing embeddings that preserve semantic similarity, so the elements from the same class are mapped to close elements in the embedding space; and elements from different classes are mapped to embeddings further apart. It's expected that averaging embeddings of elements from one class would produce good prototypes, but this is not directly modelled in the loss function.

In all experiments the network is trained using the same approach. 
Initial learning rate is fixed at $0.001$. Adam \cite{kingma2014adam} optimizer is used, as it gives reasonable results without the need to fine-tune training parameters. 
If the training stagnates and the loss on validation set doesn't decrease for a pre-defined number of epochs, the learning rate is reduced by a factor of 0.1.
We fix the size of the resulting embedding to $k=16$, to keep the amount of memory required to store a single prototype similar to the size of one FREAK descriptor.

%
%

\paragraph{Network architectures evaluation}

To find the best architecture of an embedding module, computing the function $f$ in Eq.\ref{jk:eq1}, \ref{jk:eq2}, \ref{jk:eq3} and \ref{jk:eq4}, we evaluate two types of fully-connected network architectures, named 'fat' and 'funnel'.
In 'fat' architecture all layers, but the last one, have the same number of units, equal to the input dimensionality, and the last layer has $k=16$ units. E.g. three layer 'fat' network consists of 512, 512 and 16 units.
In 'funnel' architecture the first layer has the number of units equal to the dimensionality of an input vector. Each subsequent layer has half of the units of the previous layer, and the last layer has $k=16$ units. E.g. three layer 'funnel' network consists of layers with 512, 256 and 16 units.
Each liner layer, except for the last one, is followed by SeLU \cite{klambauer2017self} non-linearity.

\begin{figure}
\centering
\label{jk:nn_search_precision_val}
\includegraphics[width=0.7\textwidth]{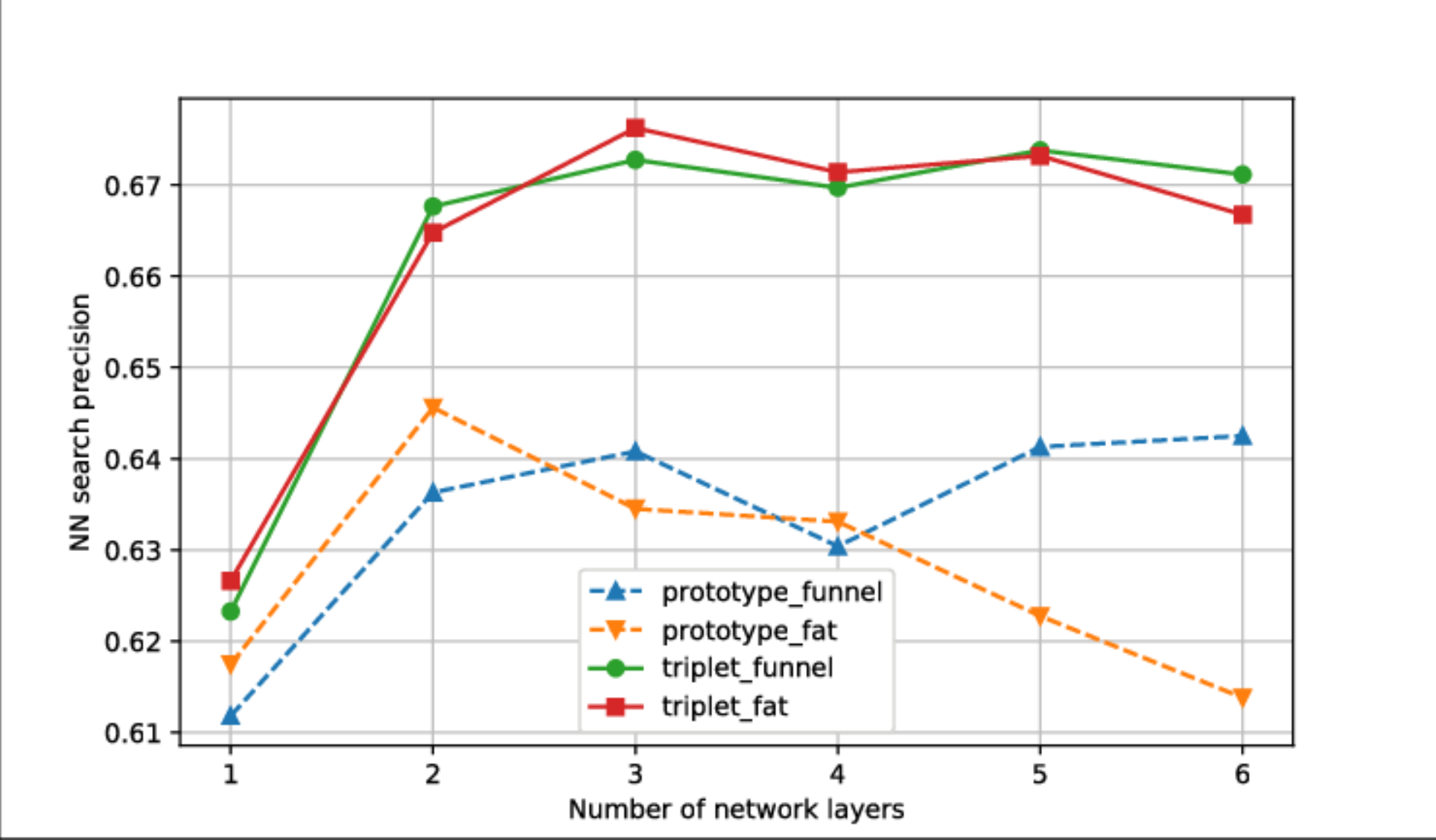}
\caption{Nearest network search precision on the validation set as a function of an embedding network architecture.
Networks trained with a triplet loss \cite{hoffer1} consistently outperform prototypical networks \cite{snell1}.
The best performance is achieved by relatively shallow three layer embedding networks trained with a triplet loss. Performance of of 'funnel' and 'fat' architectures trained with the triplet loss is very similar.}
\end{figure}

The results are presented in Fig.~\ref{jk:nn_search_precision_val}. 
Surprisingly, networks trained with a triplet loss
consistently outperform prototypical networks. 
The best performance is achieved by relatively shallow three layer embedding networks trained with a triplet loss. Performance of of 'funnel' and 'fat' architectures trained with the triplet loss is very similar. 

\begin{figure}
\centering
\label{jk:precision_out_size}
\includegraphics[width=0.7\textwidth]{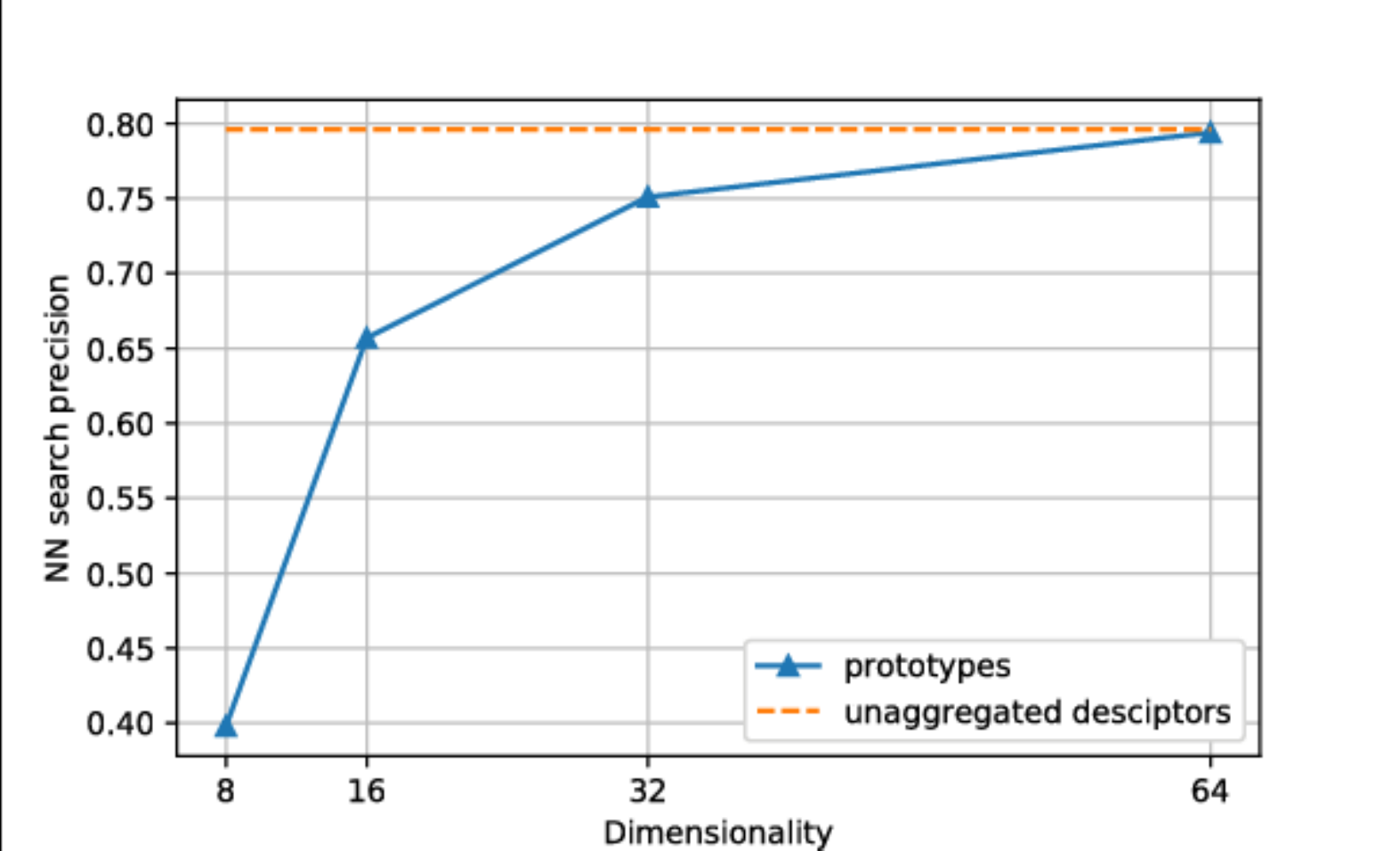}
\caption{Nearest network search precision on the test set as a function of prototype dimensionality.
The precision grows as the size of the embedding increases, reaching the performance on raw unaggregated binary descriptors for 64-dimensions.}
\end{figure}

We also investigate the dependency of the nearest neighbour search precision on the dimensionality of the prototypes constructed using our method.
We evaluated the best performing embedding network from previous experiments (three layer fully connected network with 'funnel' architecture trained with triplet loss) modifying the number of neurons in the output layer. 
This produces prototypes of different size.
The results are shown in Fig.~\ref{jk:precision_out_size}.
When dimensionality of embeddings produces by the network decreases from 16 to 8, the nearest neighbour search precision drops rapidly from $0.65$ to $0.40$.
When dimensionality increases, the search precision goes up to $0.75$ for 32 dimensions and $0.79$ for 64 dimensions.
However this is at expense of the memory required to store resulting prototypes.
Instead of storing all feature descriptors linked with a landmark, usually between 3 and 15, we can only keep one prototype.
16-dimensional prototype requires storage compared to a single FREAK descriptor. This allows reducing memory footprint a few times.
Using 32-dimensional prototypes also leads to significant memory savings, with moderate performance gap ($0.796$ for unaggregated descriptors versus $0.750$ for 32-dimensional prototypes).
When using 64-dimensional prototypes we can achieve performance comparable to using unaggregated feature descriptors. However the memory requirements, compared to 16-dimensional embeddings, quadruple.

\paragraph{Final evaluation} For the final evaluation we choose the best performing embedding network from previous experiments: three layer fully connected network with 'funnel' architecture (with 512, 256 and 16 neurons in each layer) trained with triplet loss.
The evaluation is done on a separate test set, not used during network architecture evaluation.
To benchmark the performance of the proposed method, we computed nearest neighbour search precision using alternative approaches.
Results of the evaluation are given in the Table~\ref{jk:table1}.
As expected, nearest neighbour search precision using unaggregated feature descriptors is the highest ($0.796$).
By aggregating feature descriptors linked with each landmark to 32-dimensional prototypes we can get minimally worse search precision ($0.750$).
Using 16-dimensional prototypes the search precision drops to $0.676$.
Similar search precision ($0.671$) can be achieved by a straightforward approach of computing an arithmetic mean of feature descriptors linked with each landmark and quantizing it at $0.5$ ('quantized mean').
However, as discussed in the Section \ref{jk:section:method}, such approach cannot be used in an incremental structure-from-motion processing pipeline.
For comparison we use principal component analysis (PCA) instead of the neural network to compute low dimensional, real-valued representation of binary descriptors. Fist 16 principal components are taken and their mean is used as a landmark prototype. The results are significantly worse than our method ($0.581$).
Naive approach of taking as a prototype one random feature descriptor from a list of descriptors linked with a landmark yield even lower precision ($0.526$).

\begin{table}
\label{jk:table1}
\caption{Benchmark of performance of the proposed method.
Using 32-dimensional prototypes give good performance with small gap to raw unaggregated data. 
See Section~\ref{jk:section:experimental} for description of evaluated methods.}
\begin{center}
\begin{tabular}{l@{\quad}l}
\hline
 Method & NN search precision\\
 [2pt]
\hline\rule{0pt}{12pt}
Unaggregated feature descriptors  &  0.796  \\
\textbf{32-dimensional prototypes} &  \textbf{0.750}  \\
\textbf{16-dimensional prototypes} &  \textbf{0.676}  \\
Quantized mean (512 bits) &  0.671  \\
Mean of PCA projection (16 dimensions) & 0.581 \\
Random sample (512 bits) & 0.526 \\
[2pt]
\hline
\end{tabular}
\end{center}
\end{table}

\section{Summary and future work}

In this paper we presented an effective method for aggregation of binary feature descriptors. The method allows compact scene model representation in large scale structure-from-motion applications.
This is achieved by 
by computing a single prototype representing all feature descriptors linked with a landmark. The prototype can be iteratively updated, as additional feature descriptors detected in new frames are linked with the landmark.
The size of the resulting prototype can be chosen as a trade-off between the required compression ratio and search precision.
32-dimensional prototypes offer very good performance, with small gap to unaggregated feature descriptors, and require the storage similar to the size of two 512-bit FREAK descriptors. 
Taking into account that a few feature descriptors are linked with each landmark, usually between 3 and 15 in practical applications, significant memory savings can be achieved.

Another advantage is that our method allows speeding-up of one of the key steps in the structure-from-motion pipeline, that is matching descriptors from a new frame with descriptors linked with landmarks in reconstructed 3D scene model. 
First of all, there are much less class prototypes than unaggregated feature descriptors, so the search can be performed faster.
Secondly, descriptor embeddings and class prototypes are relatively low-dimensional real valued vectors.
This allow to use very efficient approximate nearest neighbour search methods, such as FlANN~\cite{flann1}, which are work very well in real-valued spaces.

As a future work we'd like to investigate if using and encoding additional information provided by a keypoint detector (e.g. scale and orientation) can improve the search precision.
Traditionally keypoints are matched by searching using the nearest neighbour using feature descriptor values. In practice we often have additional information, such as scale and orientation, estimated by the keypoint detector. Using neural networks it should be possible to combine this information with raw binary descriptor to compute more discriminative prototypes.

\section*{Acknowledgement}
This research was supported by Google Sponsor Research Agreement under
the project "Efficient visual localization on mobile devices".

The Titan X Pascal used for this research was donated by the NVIDIA Corporation.

%
%

\end{document}